\pdfoutput=1
\documentclass[11pt]{article}

\usepackage[final]{acl}

\usepackage{times}
\usepackage{latexsym}

\usepackage[T1]{fontenc}
\usepackage[utf8]{inputenc}

\usepackage{microtype}

\usepackage{inconsolata}

\usepackage{graphicx}

\usepackage{multirow}
\usepackage{pdflscape}
\usepackage{amssymb}
\usepackage{amsmath}
\usepackage{amsfonts}
\usepackage{graphicx}
\usepackage{threeparttable}
\usepackage{booktabs}
\usepackage{xspace}
\usepackage{listings}
\definecolor{codegreen}{rgb}{0,0.6,0}
\definecolor{codegray}{rgb}{0.5,0.5,0.5}
\definecolor{codepurple}{rgb}{0.58,0,0.82}
\definecolor{backcolour}{rgb}{0.95,0.95,0.92}
\lstdefinestyle{mystyle}{
    backgroundcolor=\color{backcolour},   
    commentstyle=\color{codegreen},
    keywordstyle=\color{magenta},
    numberstyle=\tiny\color{codegray},
    stringstyle=\color{codepurple},
    basicstyle=\ttfamily\footnotesize,
    breakatwhitespace=false,         
    breaklines=true,                 
    captionpos=b,                    
    keepspaces=true,                 
    numbers=left,                    
    numbersep=5pt,                  
    showspaces=false,                
    showstringspaces=false,
    showtabs=false,                  
    tabsize=2
}
\newcommand{\lt}{\texttt{LinkTransformer}\xspace}

\begin{document}

%%%%%%%%% TITLE
\title{\lt: A Unified Package for Record Linkage with Transformer Language Models} 
\author{Abhishek Arora and Melissa Dell$^{\ast}$ \\
Harvard University, Cambridge, MA, USA \\
\normalsize{$^\ast$Corresponding author:  melissadell@fas.harvard.edu.}
}
\maketitle

%%%%%%%%% ABSTRACT
\begin{abstract}
Many computational analyses require linking information across noisy text datasets. While large language models (LLMs) offer significant promise, approximate string matching in popular statistical softwares such as R and Stata remain predominant in academic applications. These packages have simple interfaces and can be easily extended to a diversity of languages and settings, and for academic applications, ease-of-use and extensibility are essential. In contrast, packages for record linkage with LLMs require significant familiarity with deep learning frameworks and often focus on applications of commercial value in English. The open-source package \lt aims to bridge this gap by providing an end-to-end software for performing record linkage and other data cleaning tasks with transformer LLMs, treating linkage as a text retrieval problem. At its core is an off-the-shelf toolkit for applying transformer models to record linkage. \lt contains a rich repository of pre-trained models for multiple languages and supports easy integration of any transformer language model from Hugging Face or OpenAI, providing the extensibility required for many scholarly  applications. Its APIs also perform common data processing tasks, \textit{e.g.}, aggregation, noisy de-duplication, and translation-free cross-lingual linkage. \lt contains comprehensive tools for efficient model tuning, allowing for highly customized applications, and users can easily contribute their custom-trained models to its model hub to ensure reproducibility. 
Using a novel benchmark dataset geared towards academic applications, we show that \lt - with both custom models and Hugging Face or OpenAI models off-the-shelf - outperforms string matching by a wide margin.
By combining transformer LMs with intuitive APIs, \lt aims to democratize these performance gains for those who lack familiarity with deep learning frameworks. 

\end{abstract}

%%%%%%%%% BODY TEXT
\section{Introduction}

Linking information across sources is fundamental to a variety of analyses in social science, business, and government. 
A recent literature, focused on matching across e-commerce datasets, shows the promise of transformer large language models (LLMs) for improving record linkage (alternatively termed entity resolution or approximate dictionary matching). Yet these methods have not yet made widespread inroads in social science applications, with rule-based methods continuing to overwhelmingly predominate (\textit{e.g.}, see reviews by \citet{binette2022almost, abramitzky2021automated}). %A recent review ``(Almost) all of entity resolution'' in Science Advances \cite{binette2022almost} suggests that deep neural models are unlikely to be applicable to much of record linkage in structured datasets. %, and discipline specific reviews (e.g., \citet{abramitzky2021automated} for economics), likewise do not show inroads by transformer language models. 
In particular, researchers commonly employ string-based matching tools available in statistical software packages such as R or Stata. %mention any others, e.g. (\textit{e.g.}, \cite{okazaki2010simple})

In academic applications, extensibility to a diversity of human societies (historic and present) and ease of use for those not familiar with deep learning are essential. String matching algorithms in widely used statistical packages meet these requirements because they require little coding expertise and can easily be applied across different languages and settings. 
In contrast, existing tools for large language model matching require considerable technical expertise to implement. This makes sense in the context for which these models were developed - classifying and linking products for e-commerce firms, which employ data scientists - but it is a significant impediment for scholarly use.

\begin{figure*}[ht]
    \centering
    \includegraphics[width=\linewidth]{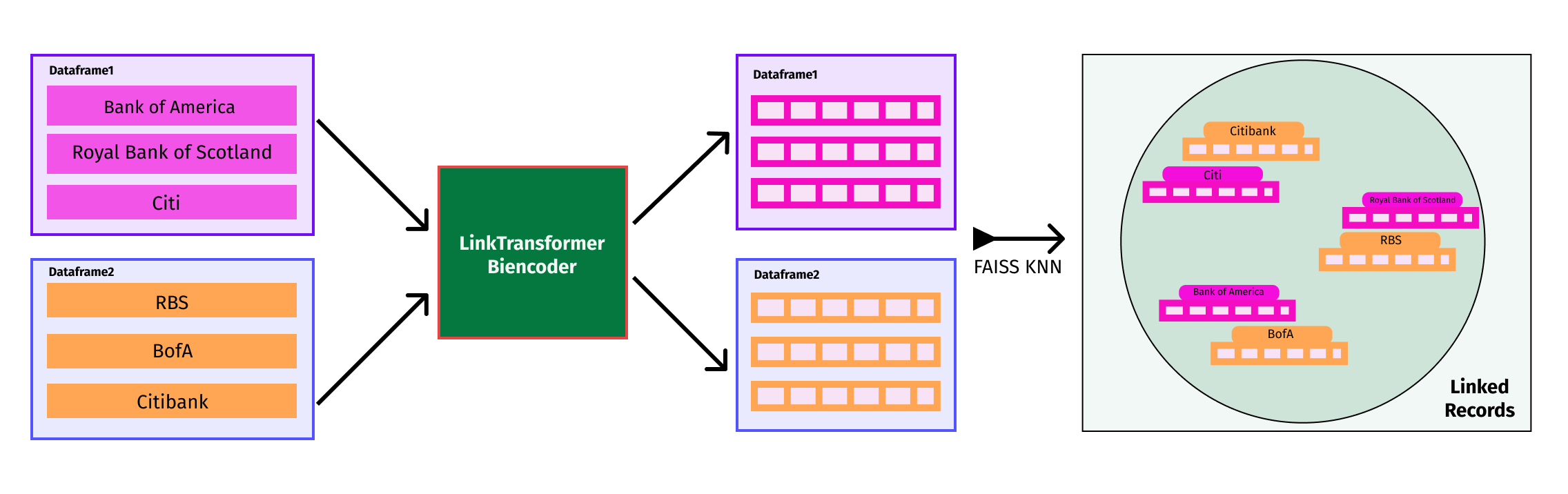}
    \caption{\textbf{Architecture.} This figure shows the \lt architecture for record linkage.} 
    \label{fig:viz}
    \vspace{-4mm}
  \end{figure*}
  
To bridge the gap between the ease-of-use of widely employed string matching packages and the power of modern LLMs, we developed \lt, a general purpose, user friendly package for record linkage with transformer LLMs. \lt treats record linkage as a text retrieval problem (See Figure 1). The API can be thought of as a drop-in replacement to popular dataframe manipulation frameworks like pandas or tools like R and Stata, catering to those who lack extensive coding experience.  

\lt integrates:
\begin{enumerate}
    \item An off-the-shelf toolkit for applying transformer models to record linkage 
    \item A rich repository of pre-trained models, supporting multiple applications and languages and evaluated on novel social science-oriented benchmarks
    \item Easy integration of any Hugging Face or OpenAI transformer LLM, for extensibility
    \item APIs to support common data processing tasks: aggregation, de-duplication, classification, and translation-free cross-lingual linkage
    \item Comprehensive tools for model tuning 
    \item Easy sharing to the \lt model hub, as reproducibility is essential for academic applications
\end{enumerate}

%The \lt model zoo currently contains English, Chinese, French, German, Japanese, Spanish, and multilingual pre-trained models. 
%These are tuned and evaluated using novel benchmark datasets that are tailored to common social science applications.
%We initialize with semantic similarity models, \textit{e.g.,} \citet{reimers2019sentence}, which have desirable properties relative to using off-the-shelf embeddings from models like RoBERTa (see Section \ref{sec:related-work}).
%We further tuned these models on a variety of linked datasets. 

While transfer learning can facilitate strong off-the-shelf performance, heterogeneity in how out-of-domain applications are from LLM training corpora - combined with settings that demand extremely high accuracy - create many scenarios where custom training may be needed. \lt makes it straightforward for users to tune their own customized models. 
%Record linkage applications are extremely diverse in their languages, time periods, and domains, which vary significantly in how out-of-domain they are from the web corpora that underlying LLMs are trained on. 

\lt performs well on challenging record linkage applications. It is equally applicable to tasks with a single field - \textit{e.g.,} linking 1940s Mexican tariff product classes across time - and applications that require concatenating an array of noisily measured fields - \textit{e.g.,} linking 1950s Japanese firms across different large-scale, noisy databases using the firm name, location, products, shareholders, and banks. This type of linkage problem would be highly convoluted with traditional string matching, as there are many noisily measured fields (\textit{e.g.,} products can be described in different ways, different subsets of managers and shareholders are listed, etc). Using \lt to automatically concatenate the information and feed it to a LLM handles these challenges with ease.

\lt has a GNU General Public License and is being actively maintained.  A demo is available at \url{https://youtu.be/hFrh8k1pukI}. More resources are available on our package website \url{https://linktransformer.github.io/}.
%It is being actively maintained, and in the next release we will add support for vision-free and multimodal linkage models, including support to import and customize any timm model \cite{rw2019timm}. When OCR errors are rampant, vision-only or aligned vision-language transformer models can improve record linkage, relative to string matching or language-only transformer linking \textit{e.g.}, \cite{yang2023quantifying, arora2023linking}.
%When OCR errors are widespread, linking using a vision only model with the original image crops or an aligned language-image model that incorporates both the noisy texts and the original image crops can improve record linakge, relative to language-only linking \cite{arora2023linking}. 
%Second, we will add GPU-FAISS \cite{johnson2019billion} backend support (currently, the package supports a CPU-FAISS backend), so that it can easily scale to extremely large datasets.

This study is organized as follows:
%The rest of the paper is organized as follows. 
Section~\ref{sec:related-work} provides an overview of related work.
The core \lt library is described in Section~\ref{sec:core-library}. 
Section~\ref{sec:use-case} evaluates \lt performance on various use cases, and
Section \ref{sec:limits} considers ethics. 

\section{Relation to the Existing Literature}
\label{sec:related-work}

%There is a large literature on record linkage spanning social science, statistics, and computer science. 
%Record linkage serves as a prerequisite for many empirical analyses, which typically involve curating data from diverse sources that may need to be linked using noisy text fields. 
The record linkage literature is sprawling - with large literatures in quantitative social science (particularly economics), statistics, computer science, and industry applications. These literatures are highly disjoint, taking very different approaches and even using different terms (record linkage, entity resolution/matching, approximate dictionary matching, etc.) to refer to the same task. 
A 2022 interdisciplinary \textit{Science Advances} review, ``(Almost) All of Entity Resolution'' \cite{binette2022almost}, concludes that deep neural models are unlikely to be applicable to record linkage using structured data. It argues that training datasets are small and there is not much to be gained from LLMs since text fields are often short. 
 Yet there is an active literature on e-commerce applications that underscores the utility of LLMs for linking structured datasets, even when text fields are short. Benchmarks in this literature (e.g., \citet{kopcke2010evaluation, das2015magellan, primpeli2019wdc}) focus on high resource commercial applications in English, such as matching electronics and software products between Amazon-Google and Walmart-Amazon listings, matching iTunes and Amazon music listings, and matching restaurants between Fodors and Zagat. Recent studies have used masked language models \cite{li2020deep, joshi2021relink, brunner2020entity, zhou2022relation}, GPT \cite{peeters2023using, tang2022generic}, or both, significantly outperforming static word embedding and other older linkage methods. 

The siloed nature of the literature is reflected in softwares. 
The main existing package for record linkage with LLMs is Ditto \cite{li2020deep}, which implements \citet{li2023effective}. It requires significant programming expertise to deploy. While this is appropriate for an e-commerce target audience - where data scientists predominate - the technical expertise required and the lack of pre-trained models targeted to multilingual social science applications has likely hindered further takeup. Moreover, most of the literature examining record linkage with LLMs poses record linkage as a classification task \cite{barlaug2021neural}, which is appropriate for the e-commerce benchmarks. However, this significantly limits extensibility, as in many social science and government applications the number of entities to be linked numbers in the millions, making it computationally infeasible to compute a softmax over all possible classes (entities). In the social sciences, string matching with statistical packages predominates.

\lt frames record linkage as a knn-retrieval task, in which the nearest neighbor for each entity in a query embedding dataset is retrieved from a key embedding dataset, using cosine similarity implemented with an FAISS backend \cite{johnson2019billion}. \lt includes functionality to tune a no-match threshold - since not all entities in the query need to have a match in the key - and allows for retrieving multiple neighbors, to accommodate many-to-many matches between the query and the key. The \lt architecture was inspired by bi-encoder applications with unstructured texts, \textit{e.g.}, passage retrieval \cite{karpukhin2020dense}, entity disambiguation \cite{wu2019scalable}, and entity co-reference resolution \cite{hsu2022contrastive}. The knn retrieval structure of \lt also supports noisy de-duplication, a closely related task that finds noisily duplicated observations within a dataset, following the methods developed in \citet{silcock2022noise}.%, who show that de-duplication using a contrastively trained bi-encoder significantly outperforms $n$-gram and locally sensitive hashing methods and is highly scalable. %Texts can be encoded using any \lt compatible transformer LLM and then the embeddings are clustered with highly scalable single linkage clustering to detect duplicates.

\lt departs from much of the literature in utilizing LLMs trained for semantic similarity, combined with a supervised contrastive loss \cite{khosla2020supervised}. Off-the-shelf LLMs such as BERT have anisotropic geometries \cite{ethayarajh2019contextual}, which makes them unsuitable off-the-shelf for metric learning problems like \lt nearest neighbor retrieval. Contrastive training for semantic similarity reduces anisotropy, improving alignment between semantically similar pairs to be linked and improving sentence embeddings \cite{wang2020understanding, reimers2019sentence}.
\lt builds closely upon the contrastively trained Sentence BERT \cite{reimers2019sentence}, whose semantic similarity library inspired many of the features in \lt.

\section{The \lt Library}
\label{sec:core-library}

\subsection{Off-the-shelf Toolkit}

At the core of \lt is an off-the-shelf toolkit that streamlines record linkage with transformer language models. 
The record linkage models enable using pre-trained or self-trained transformer models with minimal coding required. 
%The API can be thought of as a drop-in replacement to popular dataframe manipulation frameworks like pandas or tools like R and Stata, catering to those who lack extensive exposure to coding.  
Any Hugging Face or OpenAI model can be used by configuring the \texttt{model} and \texttt{openai\_key} arguments. This future-proofs the package, allowing it to take advantage of the open-source revolution that Hugging Face has pioneered.  
Here is an example of the core \textbf{merge} functionality, based on embeddings sourced from an external language model. 

\begin{lstlisting}[language=python]
#pip install linktransformer 
import linktransformer as lt
df1 = pd.read_csv("df1.csv")
df1 = pd.read_csv("df2.csv")
df_matched = lt.merge(df2, df1, merge_type='1:m', on=["Varname"], model="sentence-transformers/all-MiniLM-L6-v2", openai_key=None)
\end{lstlisting}

We recommend that users new to LLMs deploy the package using a cloud service optimized for deep learning to avoid the need to resolve dependencies, and our tutorials use Colab.

In addition to supporting Hugging Face and OpenAI models, \lt provides pre-trained models, currently encompassing six languages (English, Chinese, French, German, Japanese, and Spanish) plus a multilingual model. These models are trained and evaluated using novel datasets that reflect common record linkage tasks in quantitative social science:
\begin{enumerate}
    \item \textbf{Firm aliases}: these are drawn from Wikidata for 6 languages. Firm alias models learn to recognize the different ways that firm names are written and abbreviated.
    \item \textbf{Homogenized industry and product names:} These are drawn from the United Nations economic classification schedules (International Standard Industrial Classification, Standard International Trade Classification, and Central Product Classification), that map different country classifications to an international standard. We include models trained on these for 3 official UN languages.%\footnote{https://www.un.org/en/our-work/official-languages} 
    \item \textbf{Historical datasets}: linked product-level Mexican tariff schedules from 1947 and 1948, and a dataset linking 1950s Japanese firms across noisy text databases. Historical data are central to better understanding economic and social processes; for example, these datasets could be used to elucidate the political determinants of tariff policy or the role of supply chain linkages in Japan's spectacular 20th century growth performance.
\end{enumerate}

We also provide models for the standard industry benchmarks.

We name these models with a semantic syntax:  \verb+{org_name}/lt-{data}-{task}-{lang}+.
Each model has a detailed model card, %describing model usage for inference and fine-tuning, training dataset descriptions, and other training details, along 
with the appropriate tags for model discovery. Additionally, we provide a high-level interface to download the right model by task through a wrapper that retrieves the best model for a task chosen by the user.

\lt makes no compromise in scalability. All functions are vectorized wherever possible and the vector similarity search underlying knn retrieval is accelerated by an FAISS \cite{johnson2019billion} backend that can easily be extended to perform retrieval on GPUs with billion-scale datasets. We also allow ``blocking'' - running knn-search only within ``blocks'' that can be defined by the \texttt{blocking\_vars} argument. %Consider an example : 

% \begin{lstlisting}[language=python]
% #Load data frame
% df1 = pd.read_csv("df1.csv")
% df1 = pd.read_csv("df2.csv")
% df_matched = lt.merge(df2, df1, merge_type='1:m', on=["Varname"], model="sentence-transformers/all-MiniLM-L6-v2", openai_key=None,blocking_vars=["blockingvar"])
% \end{lstlisting}

Record linkage frequently requires matching databases on multiple noisily measured keys. \lt allows a list of as many variables as needed in the "on" argument. %This example fuzzily matches two dataframes with company names and their product information, 
% \begin{lstlisting}[language=python]
% #Load data frame
% df1 = pd.read_csv("df1.csv")
% df1 = pd.read_csv("df2.csv")
% df_matched = lt.merge(df2, df1, merge_type='1:m', on= ["CompanyName","ProductDescription"], model="sentence-transformers/all-MiniLM-L6-v2", openai_key=None,blocking_vars=["blockingvar"])
% \end{lstlisting}
The merge keys specified by the \texttt{on} variable are serialized by concatenating them with a $<SEP>$ token, which is based on the underlying tokenizer of the selected base language model. %This ensures that the serialization takes advantage of a token already introduced in training, and the process is agnostic to the choice of the model.% as long as it has a tokenizer module supported by the \textit{Autotokenizer} class of Hugging Face - which most transformer based LLMs do. 
Since we have designed the API around dataframes - due to their familiarity amongst users of R, Stata or Excel - all import/export formats are supported.

The \lt API supports a plethora of other features that are frequently integrated into data analysis pipelines. These include: 

\textbf{Aggregation:}
Data processing often requires the aggregation of fine descriptions into coarser categories, that are consistent across datasets and time or facilitate interpretation. This problem can be thought of as a merge between finer categories and coarser ones, where \lt classifies the finer categories by means of finding their nearest coarser neighbor(s). \texttt{lt.aggregate\_rows} performs this task, with a similar syntax to the main record linkage API.  

% \begin{lstlisting}[language=python]
% df_coarse = pd.read_csv(os.path.join("coarse.csv"))
% df_fine = pd.read_csv(os.path.join("fine.csv"))
% df_aggregated = lt.aggregate_rows(df_fine, df_coarse, model="sentence-transformers/all-mpnet-base-v2", left_on="Fine Category Name", right_on="Coarse Category Name")

% \end{lstlisting}

\textbf{Deduplication}:
Text datasets can contain noisy duplicates.
Popular libraries like dedupe \cite{gregg2022dedupe} only support deduplication using metrics that most closely resemble edit distance. \lt allows for semantic deduplication with a single, intuitive function call. 

\begin{lstlisting}[language=python]
df_dedup=lt.dedup_rows(df,on="CompanyName",model="sentence-transformers/all-MiniLM-L6-v2",
    cluster_params= {'threshold': 0.7})
\end{lstlisting}

\lt de-duplication clusters embeddings under the hood, with embeddings in the same cluster classified as duplicates. \lt supports SLINK, DBSCAN, HDBSCAN, and agglomerative clustering. %For users who are interested in the cluster assignments, we also provide a similarly convenient function for clustering. 

% \begin{lstlisting}[language=python]
% df=pd.read_csv("df1.csv"))
% df_clusters=lt.cluster_rows(df,on="CompanyName",model="sentence-transformers/all-MiniLM-L6-v2",
%     cluster_params= {'threshold': 0.7})
% \end{lstlisting}

\textbf{Cross-lingual linkage}:
Analyses spanning multiple countries often require cross-lingual linkage. Machine translation followed by string matching methods tend to perform very poorly, necessitating costly hand linking. \lt users can bypass translation by using multilingual transformer models. %, some of which are available in our model zoo.
% Consider an example where we link product descriptions in French to those in English: 
% \begin{lstlisting}[language=python]
% df_french=pd.read_csv("df1.csv"))
% df_english=pd.read_csv("df2.csv"))
% df_lm_matched = lt.merge(df_french, df_english, merge_type='1:m', left_on="Libelle du Produit", right_on="Product Label", model="distiluse-base-multilingual-cased-v1")
% \end{lstlisting}

\textbf{Text Classification:}  While Hugging Face provides an accessible API, text classification can still be challenging for users who haven't been exposed to NLP libraries. Our API requires only one line of code to use a classification model on Hugging Face or the ChatGPT (3.5 and 4)  API to classify texts. %As before, the wrapper can handle multiple columns of inputs by concatenating the \texttt{on} columns with a $<SEP>$ token.

%\begin{lstlisting}[language=python]
%df_clf = lt.classify_rows(df, on=[ "headline","article"], model="gpt-3.5-turbo", %num_labels=2, openai_key="sk-...", openai_topic="protests")
%\end{lstlisting}

Notebooks and tutorials outline the use of these functionalities on toy datasets.\footnote{\url{https://linktransformer.github.io/}} We also have a tutorial to help those who are less familiar with language models to select ones that fit their use case. 
More detailed information and additional features can be found in the online documentation.\footnote{\url{https://github.com/dell-research-harvard/linktransformer}}

\subsection{Customized Model Training}
\label{sec:training}

\subsubsection*{Record Linkage}

Record linkage tasks are highly diverse and may demand very high accuracy; hence, fine-tuning on target datasets may be necessary. \lt supports easy model training, which can be initialized using any Hugging Face transformer model.

Training data are expected in a \textit{pandas} data frame, removing entry barriers for the typical social science user. 
A data frame can include only positive labeled examples (linked observations) as inputs, in which case the model is evaluated using an information retrieval evaluator that measures top-1 retrieval accuracy. Alternately, it can take a list of both positive and negative pairs, in which case the model is evaluated using a binary classification objective. 

Only the most important arguments are exposed and the rest have reasonable defaults which can be tweaked by more advanced users. Additionally, \lt supports logging of a training run on Weights and Biases \cite{wandb}.

\begin{lstlisting}[language=python]
best_model_path=lt.train_model(
        model_path="hf-path-model", 
        data="df1.csv",
        left_col_names=["left_var"],
        right_col_names=['right_var'],
        label_col_name=None,
        log_wandb=False,
        training_args={"num_epochs": 1})
\end{lstlisting}

Default training expects positive pairs. A simple argument that specifies \texttt{label\_col\_name} switches the dataset format and model evaluation to adapt to positive and negative labels. To make this extensible to most record linkage use-cases, the model can also be trained on a dataset of cluster ids and texts by simply specifying \texttt{clus\_id\_col\_name} and \texttt{clus\_text\_col\_names}. 

\lt is sufficiently sample efficient that most models in the model zoo were trained with a student Google Colab account, an integral feature since the vast majority of potential users have constrained compute budgets. 

\subsubsection*{Classification}
We added classification at the request of \lt users.
Users can train custom models with a single line of code, using training data in the form of a data frame. They simply specify the \texttt{on} columns containing the text and a column for annotations, \texttt{label\_col\_name}. We have helpful guides on our website to allow users to effectively tune hyperparameters. 

%\begin{lstlisting}[language=python]
%best_model_path,best_metric,label_map= %lt.train_clf_model(data=train_data,model="distilroberta-base",
%    on=["headline","byline","text"],label_col_name="label",
%    lr=2e-5,batch_size=8,
%    training_args={"num_train_epochs":10})
%\end{lstlisting}

Since this function wraps around the \textit{Trainer} class from Hugging Face, it can make use of multiple GPUs. \texttt{training\_args} allow an advanced user to fully customize the \textit{Trainer} by providing arguments with the same format as Hugging Face's \textit{TrainingArguments} class.

\subsection{User Contributions}
\lt aims to promote reusability and reproducibility, which are central to academic applications. 
End-users can upload their self-trained models to the \lt Hugging Face hub with a simple 
 \texttt{model.save\_to\_hub}
command. Whenever a model is saved, a model card is automatically generated that follows best practices outlined in Hugging Face's Model Card Guidebook. %\footnote{https://Hugging Face.co/docs/hub/model-card-guidebook}. 
%This process adds a pipeline-tag, supported language(s) (given the base model), and other tags to the model card's header that facilitate model discovery by other Hugging Face users. Moreover, the automatically generated card contains instructions on how to use the model for record linkage and model-specific architecture and training details.% in the interest of reproducibility. 

%\begin{lstlisting}[language=python]
%###Load the best model
%model=lt.load_model(best_model_path)

%model.save_to_hub(repo_name = "linktransformer-model-name", ##Write model name here
%                    organization= "ORG_NAME",
%                    commit_message = "Add new LinkTransformer model.",
 %                   replace_model_card = True,
 %                   )
%\end{lstlisting}

\section{Applications}
\label{sec:use-case}

\begin{table}[ht]
    \resizebox{0.98\linewidth}{!}{
    \begin{threeparttable}
    \centering
\begin{tabular}{lcccc}
        \toprule
    \textbf{Model} & \textbf{Edit Distance} & \textbf{SBERT} & \textbf{LT} & \textbf{OpenAI} \\
    \midrule
    \multicolumn{5}{l}{\textit{Panel A: Company Linkage}} \\
    lt-wikidata-comp-fr & 0.43 & 0.74 & \textbf{0.81} & 0.75 \\
    lt-wikidata-comp-ja & 0.51 & 0.61 & \textbf{0.70} & 0.63 \\
    lt-wikidata-comp-zh & 0.66 & 0.77 & \textbf{0.83} & 0.82 \\
    lt-wikidata-comp-de & 0.51 & 0.66 & \textbf{0.76} & 0.71 \\
    lt-wikidata-comp-es & 0.62 & 0.68 & 0.75 & \textbf{0.82} \\
    lt-wikidata-comp-en & 0.36 & 0.60 & \textbf{0.70} & 0.64 \\
    lt-wikidata-comp-multi & 0.55 & 0.69 & \textbf{0.83} & 0.77 \\
    lt-wikidata-comp-prod-ind-ja & 0.48 & 0.97 & \textbf{0.99} & 0.98 \\
    \multicolumn{5}{l}{\textit{Panel B: Fine Product Linkage}} \\
    lt-un-data-fine-fine-en & 0.64 & 0.82 & \textbf{0.87} & 0.84 \\
    lt-un-data-fine-fine-es & 0.42 & 0.68 & \textbf{0.80} & 0.72 \\
    lt-un-data-fine-fine-fr & 0.45 & 0.71 & \textbf{0.75} & 0.72 \\
    lt-un-data-fine-fine-multi & 0.54 & 0.79 & \textbf{0.84} & 0.77 \\
    \multicolumn{5}{l}{\textit{Panel C: Product to Industry Linkage}} \\
    lt-un-data-fine-industry-en & 0.18 & 0.81 & \textbf{0.80} & 0.73 \\
    lt-un-data-fine-industry-es & 0.18 & 0.67 & \textbf{0.72} & 0.64 \\
    lt-un-data-fine-industry-fr & 0.14 & 0.56 & \textbf{0.72} & 0.55 \\
    lt-un-data-fine-industry-multi & 0.10 & 0.69 & \textbf{0.78} & 0.75 \\
    \multicolumn{5}{l}{\textit{Panel D: Product Aggregation}} \\
    lt-un-data-fine-coarse-en & 0.27 & 0.76 & 0.85 & \textbf{0.86 }\\
    lt-un-data-fine-coarse-es & 0.24 & 0.75 & \textbf{0.80} & 0.7 \\
    lt-un-data-fine-coarse-fr & 0.24 & 0.74 & \textbf{0.77} & 0.69 \\
    lt-un-data-fine-coarse-multi & 0.22 & 0.6 & \textbf{0.64} & 0.62 \\
    \bottomrule
\end{tabular}
\end{threeparttable}}
    \caption{Performance of various embedding models, measured by top-1 retrieval accuracy. \textit{Company linkage} links company aliases together,  \textit{Fine Product Linkage} links products from different product classifications together, \textit{Product to Industry Linkage} links products to their industry classifications, and \textit{Product Aggregation} links a fine product to its coarser product classification. \textit{LT} gives the performance of the trained LinkTransformer model. \textit{Edit Distance} gives linkage accuracy when using Levenshtein distance, and \textit{SBERT} when using semantic similarity models off-the-shelf (See Table \ref{tab:languages_models}). \textit{OpenAI} gives the best linkage performance when using embeddings from OpenAI embedding models.}
     \label{tab:performance_metrics}
\end{table}

The LLMs in the \lt model zoo excel at a variety of tasks.
Table \ref{tab:performance_metrics} evaluates performance linking Wikidata firm aliases (panel A), linking product descriptions from different countries' classification schemes (panel B), linking products to their industries (panel C), and aggregating fine product descriptions to coarser descriptions (panel D). It compares the accuracy of Levenshtein edit distance matching \cite{levenshtein1966binary}, popular off-the-shelf semantic similarity models from Hugging Face (see Appendix Table \ref{tab:languages_models} for a listing of models used), OpenAI embeddings (the better of \texttt{text-embedding-3-small} and \texttt{text-embedding-3-large}, which outperformed Ada embeddings), and \lt tuned models. The supplementary materials describe the models and training datasets in detail. 

As expected, custom-tuned models typically achieve the best performance, with off-the-shelf models still outperforming edit distance matching, typically by a wide margin.
The custom-trained models are often plausibly achieving human-level accuracy, as cases that they get wrong are often impossible to resolve from the information provided, \textit{e.g.,} in cases where a firm is referred to by two completely disparate acronyms.

Second, we examine historical applications, which are central to understanding long-run phenomena like economic growth or social mobility and typically lack unique identifiers for linkage. First, we link two tariff schedules published by the Mexican government in the 1940s \cite{MxCrosswalk1948}. Tariffs were applied at an extremely disaggregated product level and each of the many thousands of products in the tariff schedule is identified only by a text description, which can change each time the tariff schedule is updated. Around 2,000 products map to different descriptions across the schedules in a rare crosswalk published by the government (typically, homogenized crosswalks do not exist). 
We link the tariff schedules using an off-the-shelf semantic similarity model, as well as a model tuned on the in-domain historical data and Open AI embeddings. All transformer models widely outperform edit distance. While there are considerable debates on the role that trade policies have played in long-run development, empirical evidence is limited largely due to the considerable challenges of homogenizing tariff schedules across time. Language model linking offers the opportunitiy to bring novel quantitative evidence to this important question.

\begin{table}[t]
  \resizebox{\linewidth}{!}{
    \begin{threeparttable}
    \centering
    \begin{tabular}{lccccc}
        \toprule
        \textbf{Dataset} & \textbf{Semantic} & \textbf{Fine}  & \textbf{Edit} & \textbf{OpenAI} & \textbf{LT} \\
        & \textbf{Sim} & \textbf{Tuned} & \textbf{Distance} &\textbf{ADA}   &\textbf{UN/Wiki Model} \\
        \midrule
        mexicantrade4748 & 0.75 & \textbf{0.88} & 0.70 & 0.83 & 0.80 \\
        historicjapan & 0.69 & \textbf{0.91} & 0.27 & 0.86 & 0.74 \\
        \bottomrule
    \end{tabular}
  \end{threeparttable}}
  \caption{Historical Linking. We examine the base semantic similarity model off-the-shelf, a fine-tuned \lt version, Levenshtein edit distance on the tariff description or company name, OpenAI embeddings and a pre-trained LinkTransformer model. The table reports top-1 accuracy. }
      \label{tab:historical}
\end{table}

We also link firms across two different 1950s publications created by different Japanese credit bureaus \cite{pr, teikoku}. One has around 7,000 firms and the other has around 70,000, including many small firms. %We recognized the complex table layouts using a Mask R-CNN \cite{he2017mask} model custom-trained with Layout Parser \cite{shen2021layoutparser} and digitized the texts with custom OCR \cite{carlson2023}. 
Firm names can be written differently across publications and there are many duplicated or similar firm names. 
We concatenate information on the firm's name, prefecture, major products, shareholders, and banks. These variables contain OCR noise and the information included varies, \textit{e.g.} in terms of how a firm's products are described, which shareholders are included, etc. This makes rule-based methods quite brittle, whereas the custom-tuned model links 91\% of firms correctly.

In the supplemental materials, we examine the various e-commerce and industry benchmarks that prevail in this literature. We use the same training procedure for each benchmark, to avoid overfitting, which is often not the case in the literature. We have generally comparable performance, sometimes outperformed by other models (that could be used with \lt if on Hugging Face) and sometimes outperforming other models.

When OCR errors are severe, too much information may have been destroyed to achieve the desired accuracy with the garbled texts. A multimodal matching framework \cite{arora2023linking} that uses aligned language and vision transformers to incorporate the original image crops or a matching framework that incorporates character visual similarity \cite{yang2023quantifying} - as OCR confuses visually similar characters - may be required. Vision and multimodal linking support will be incorporated into future releases of \lt. 

%% Does not count against page limit
\section{Ethics Statement}
\label{sec:limits}

\lt is ethically sound. It is built using public domain training data. Because it is built upon transformer language models, it will not be suitable for lower resource languages that lack pre-trained LLMs. However, it can utilize any Hugging Face or OpenAI embedding model and hence will be extensible as the low-resource transformer literature expands to lower resource settings, as long as relevant embedding models are posted on Hugging Face or made available commercially by OpenAI. 

\bibliography{cites}

\begin{thebibliography}{30}
\providecommand{\natexlab}[1]{#1}

\bibitem[{Abramitzky et~al.(2021)Abramitzky, Boustan, Eriksson, Feigenbaum, and
  P{\'e}rez}]{abramitzky2021automated}
Ran Abramitzky, Leah Boustan, Katherine Eriksson, James Feigenbaum, and
  Santiago P{\'e}rez. 2021.
\newblock Automated linking of historical data.
\newblock \emph{Journal of Economic Literature}, 59(3):865--918.

\bibitem[{Arora et~al.(2023)Arora, Yang, Jheng, and Dell}]{arora2023linking}
Abhishek Arora, Xinmei Yang, Shao~Yu Jheng, and Melissa Dell. 2023.
\newblock Linking representations with multimodal contrastive learning.
\newblock \emph{arXiv preprint arXiv:2304.03464}.

\bibitem[{Barlaug and Gulla(2021)}]{barlaug2021neural}
Nils Barlaug and Jon~Atle Gulla. 2021.
\newblock Neural networks for entity matching: A survey.
\newblock \emph{ACM Transactions on Knowledge Discovery from Data (TKDD)},
  15(3):1--37.

\bibitem[{Biewald(2020)}]{wandb}
Lukas Biewald. 2020.
\newblock \href {https://www.wandb.com/} {Experiment tracking with weights and
  biases}.
\newblock Software available from wandb.com.

\bibitem[{Binette and Steorts(2022)}]{binette2022almost}
Olivier Binette and Rebecca~C Steorts. 2022.
\newblock (almost) all of entity resolution.
\newblock \emph{Science Advances}, 8(12):eabi8021.

\bibitem[{Brunner and Stockinger(2020)}]{brunner2020entity}
Ursin Brunner and Kurt Stockinger. 2020.
\newblock Entity matching with transformer architectures-a step forward in data
  integration.
\newblock In \emph{23rd International Conference on Extending Database
  Technology, Copenhagen, 30 March-2 April 2020}, pages 463--473.
  OpenProceedings.

\bibitem[{Das et~al.(2015)Das, Doan, Psgc, Konda, Govind, and
  Paulsen}]{das2015magellan}
Sanjib Das, A~Doan, C~Gokhale Psgc, Pradap Konda, Yash Govind, and Derek
  Paulsen. 2015.
\newblock \href {https://sites. google. com/site/anhaidgroup/useful-stuff/data}
  {The magellan data repository}.

\bibitem[{Ethayarajh(2019)}]{ethayarajh2019contextual}
Kawin Ethayarajh. 2019.
\newblock How contextual are contextualized word representations? comparing the
  geometry of bert, elmo, and gpt-2 embeddings.
\newblock \emph{arXiv preprint arXiv:1909.00512}.

\bibitem[{Gregg and Eder(2022)}]{gregg2022dedupe}
Forest Gregg and Derek Eder. 2022.
\newblock \href {https://github.com/dedupeio/dedupe} {dedupe}.

\bibitem[{Hsu and Horwood(2022)}]{hsu2022contrastive}
Benjamin Hsu and Graham Horwood. 2022.
\newblock Contrastive representation learning for cross-document coreference
  resolution of events and entities.
\newblock \emph{arXiv preprint arXiv:2205.11438}.

\bibitem[{{Jinji Koshinjo}(1954)}]{pr}
{Jinji Koshinjo}. 1954.
\newblock \emph{Nihon shokuinrokj}.
\newblock Jinji Koshinjo.

\bibitem[{Johnson et~al.(2019)Johnson, Douze, and
  J{\'e}gou}]{johnson2019billion}
Jeff Johnson, Matthijs Douze, and Herv{\'e} J{\'e}gou. 2019.
\newblock Billion-scale similarity search with gpus.
\newblock \emph{IEEE Transactions on Big Data}, 7(3):535--547.

\bibitem[{Joshi et~al.(2021)Joshi, Somani, and Roy}]{joshi2021relink}
Salil~Rajeev Joshi, Arpan Somani, and Shourya Roy. 2021.
\newblock Relink: Complete-link industrial record linkage over hybrid feature
  spaces.
\newblock In \emph{2021 IEEE 37th International Conference on Data Engineering
  (ICDE)}, pages 2625--2636. IEEE.

\bibitem[{Karpukhin et~al.(2020)Karpukhin, O{\u{g}}uz, Min, Lewis, Wu, Edunov,
  Chen, and Yih}]{karpukhin2020dense}
Vladimir Karpukhin, Barlas O{\u{g}}uz, Sewon Min, Patrick Lewis, Ledell Wu,
  Sergey Edunov, Danqi Chen, and Wen-tau Yih. 2020.
\newblock Dense passage retrieval for open-domain question answering.
\newblock \emph{arXiv preprint arXiv:2004.04906}.

\bibitem[{Khosla et~al.(2020)Khosla, Teterwak, Wang, Sarna, Tian, Isola,
  Maschinot, Liu, and Krishnan}]{khosla2020supervised}
Prannay Khosla, Piotr Teterwak, Chen Wang, Aaron Sarna, Yonglong Tian, Phillip
  Isola, Aaron Maschinot, Ce~Liu, and Dilip Krishnan. 2020.
\newblock Supervised contrastive learning.
\newblock \emph{arXiv preprint arXiv:2004.11362}.

\bibitem[{K{\"o}pcke et~al.(2010)K{\"o}pcke, Thor, and
  Rahm}]{kopcke2010evaluation}
Hanna K{\"o}pcke, Andreas Thor, and Erhard Rahm. 2010.
\newblock Evaluation of entity resolution approaches on real-world match
  problems.
\newblock \emph{Proceedings of the VLDB Endowment}, 3(1-2):484--493.

\bibitem[{Levenshtein et~al.(1966)}]{levenshtein1966binary}
Vladimir~I Levenshtein et~al. 1966.
\newblock Binary codes capable of correcting deletions, insertions, and
  reversals.
\newblock In \emph{Soviet physics doklady}, volume~10, pages 707--710. Soviet
  Union.

\bibitem[{Li et~al.(2023)Li, Li, Suhara, Doan, and Tan}]{li2023effective}
Yuliang Li, Jinfeng Li, Yoshi Suhara, AnHai Doan, and Wang-Chiew Tan. 2023.
\newblock Effective entity matching with transformers.
\newblock \emph{The VLDB Journal}, pages 1--21.

\bibitem[{Li et~al.(2020)Li, Li, Suhara, Doan, and Tan}]{li2020deep}
Yuliang Li, Jinfeng Li, Yoshihiko Suhara, AnHai Doan, and Wang-Chiew Tan. 2020.
\newblock Deep entity matching with pre-trained language models.
\newblock \emph{arXiv preprint arXiv:2004.00584}.

\bibitem[{Peeters and Bizer(2023)}]{peeters2023using}
Ralph Peeters and Christian Bizer. 2023.
\newblock Using chatgpt for entity matching.
\newblock \emph{arXiv preprint arXiv:2305.03423}.

\bibitem[{Primpeli et~al.(2019)Primpeli, Peeters, and Bizer}]{primpeli2019wdc}
Anna Primpeli, Ralph Peeters, and Christian Bizer. 2019.
\newblock The wdc training dataset and gold standard for large-scale product
  matching.
\newblock In \emph{Companion Proceedings of The 2019 World Wide Web
  Conference}, pages 381--386.

\bibitem[{Reimers and Gurevych(2019)}]{reimers2019sentence}
Nils Reimers and Iryna Gurevych. 2019.
\newblock Sentence-bert: Sentence embeddings using siamese bert-networks.
\newblock \emph{arXiv preprint arXiv:1908.10084}.

\bibitem[{{Secretaria de Economía de Mexico}(1948)}]{MxCrosswalk1948}
{Secretaria de Economía de Mexico}. 1948.
\newblock Ajuste de las fracciones de la tarifa arancelaria que rigieron hasta
  el año de 1947 con las de la tarifa que entró en vigor por decreto de fecha
  13 de diciembre del mismo año y se consideraron a partir de 1948.
\newblock In \emph{Anuario Estadístico del Comercio Exterior de los Estados
  Unidos Mexicanos}. Gobierno de Mexico.

\bibitem[{Silcock et~al.(2023)Silcock, D'Amico-Wong, Yang, and
  Dell}]{silcock2022noise}
Emily Silcock, Luca D'Amico-Wong, Jinglin Yang, and Melissa Dell. 2023.
\newblock Noise-robust de-duplication at scale.
\newblock \emph{International Conference on Learning Representations}.

\bibitem[{Tang et~al.(2022)Tang, Zuo, Cao, and Madden}]{tang2022generic}
Jiawei Tang, Yifei Zuo, Lei Cao, and Samuel Madden. 2022.
\newblock Generic entity resolution models.
\newblock In \emph{NeurIPS 2022 First Table Representation Workshop}.

\bibitem[{{Teikoku Koshinjo}(1957)}]{teikoku}
{Teikoku Koshinjo}. 1957.
\newblock \emph{Teikoku Ginko Kaisha Yoroku}.
\newblock Teikoku Koshinjo.

\bibitem[{Wang and Isola(2020)}]{wang2020understanding}
Tongzhou Wang and Phillip Isola. 2020.
\newblock Understanding contrastive representation learning through alignment
  and uniformity on the hypersphere.
\newblock In \emph{International Conference on Machine Learning}, volume 119,
  pages 9929--9939. PMLR.

\bibitem[{Wu et~al.(2019)Wu, Petroni, Josifoski, Riedel, and
  Zettlemoyer}]{wu2019scalable}
Ledell Wu, Fabio Petroni, Martin Josifoski, Sebastian Riedel, and Luke
  Zettlemoyer. 2019.
\newblock Scalable zero-shot entity linking with dense entity retrieval.
\newblock \emph{arXiv preprint arXiv:1911.03814}.

\bibitem[{Yang et~al.(2023)Yang, Arora, Jheng, and Dell}]{yang2023quantifying}
Xinmei Yang, Abhishek Arora, Shao-Yu Jheng, and Melissa Dell. 2023.
\newblock Quantifying character similarity with vision transformers.
\newblock \emph{arXiv preprint arXiv:2305.14672}.

\bibitem[{Zhou et~al.(2022)Zhou, Huang, Li, and Lai}]{zhou2022relation}
Huchen Zhou, Wenfeng Huang, Mohan Li, and Yulin Lai. 2022.
\newblock Relation-aware entity matching using sentence-bert.
\newblock \emph{Computers, Materials \& Continua}, 71(1).

\end{thebibliography}

\clearpage

% \onecolumn

% \begin{center}
%     \section*{Supplementary Materials}

% \end{center}

\appendix

\section{Supplementary Materials}

\setcounter{table}{0}
\renewcommand{\thetable}{A-\arabic{table}} % Setting the table number output to letters 
\setcounter{figure}{0}
\renewcommand{\thefigure}{A-\arabic{figure}} % Setting the figure number output to letters 

\subsection{Training and other details}

\lt models use AdamW as the optimizer with a linear schedule with a 100\% warm-up with 2e-6 as the max learning rate. We use a batch size of 64 for models trained with Wikidata (companies) and UN data (products). For industry benchmarks, we used a batch size of 128. We trained the models for 150 epochs for industrial benchmarks and 100 epochs for UN/Wikidata/Historic applications. We used Supervised Contrastive loss \cite{khosla2020supervised}   and Online Contrastive loss with default hyperparameters as the training objective depending upon the structure of the training dataset (as specified in Table \ref{tab:model-losses}). The implementation for the losses was based on the implementation shared on the sentence-transformers repository \cite{reimers2019sentence}. 

\lt uses IndexFlatIP from FAISS \cite{johnson2019billion} as the index of choice, allowing an exhaustive search to get $k$ nearest neighbours. We use the inner-product as the metric. All embeddings from the encoders are L2-normalized such that the distances (inner-products) given by the FAISS indices are equivalent to cosine similarity. 

Code to replicate the below tables and train the models is available on our repository, which also contains links to our training data.

\subsection{Datasets and Results}

Table \ref{tab:languages_models} lists the base sentence transformer models that we used to initialize the custom \texttt{LinkTransformer} models. 
Table \ref{tab:description} describes the datasets used for training the \lt model zoo. They are drawn from multilingual UN product and industry concordances, Wikidata company aliases, a 1948 Mexican government concordance between tariff schedules \cite{MxCrosswalk1948}, and a hand-linked dataset between two 1950s Japanese firm-level datasets collected by credit bureaus, one containing around 7,000 firms and the other around 70,000 \cite{teikoku, pr}.
Table \ref{tab:performance_splits} describes the train-val-test splits for each of these datasets. 
Table \ref{tab:industry_benchmarks} reports results on standard industry and e-commerce benchmarks for record linkage.
\begin{table}[t]
    \centering
        \resizebox{\linewidth}{!}{

    \begin{threeparttable}
    \begin{tabular}{lc}
        \toprule
        \textbf{Language} & \textbf{Base Model} \\
        \midrule
        English & sentence-transformers/multi-qa-mpnet-base-dot-v1 \\
        Japanese & oshizo/sbert-jsnli-luke-japanese-base-lite \\
        French & dangvantuan/sentence-camembert-large \\
        Chinese & DMetaSoul/sbert-chinese-qmc-domain-v1 \\
        Spanish & hiiamsid/sentence\_similarity\_spanish\_es \\
        German & Sahajtomar/German-semantic \\
        Multilingual & sentence-transformers/paraphrase-multilingual-manet-base-v2 \\
        \bottomrule
    \end{tabular}
    \end{threeparttable}
    }
    \caption{We used the above sentence-transformers models for different languages as base models to train \lt models. They were selected from the Hugging Face model hub and the names correspond to the repo names on the Hub.}
    \label{tab:languages_models}
\end{table}
\begin{table}[htbp]
    \centering
    \resizebox{\linewidth}{!}{
    \begin{threeparttable}
    \begin{tabular}{lcl}
        \toprule
        \textbf{Model} & \textbf{Training Data} \\
        \midrule
        lt-wikidata-comp-en  & Wikidata English-language  \\
        & company names.  \\
        lt-wikidata-comp-fr  &  Wikidata French-language  \\
        & company names.\\
        lt-wikidata-comp-de  &  Wikidata German-language  \\
        & company names.\\
        lt-wikidata-comp-ja  &  Wikidata Japanese-language  \\
        & company names.\\
        lt-wikidata-comp-zh  &  Wikidata Chinese-language  \\
        & company names.\\
        lt-wikidata-comp-es  &  Wikidata Spanish-language  \\
        & company names.\\
        lt-wikidata-comp-multi & Wikidata multilingual company \\
        & names (en, fr, es, de, ja, zh). \\
        lt-wikidata-comp-prod-ind-ja & Wikidata Japanese-language \\
        & company names and industries. \\
        lt-un-data-fine-fine-en & UN fine-level product data \\
        & in English. \\
        lt-un-data-fine-coarse-en & UN coarse-level product data \\
        & in English. \\
        lt-un-data-fine-industry-en & UN product data linked \\
        & to industries in English. \\
        lt-un-data-fine-fine-es & UN fine-level product data \\
        & in Spanish. \\
        lt-un-data-fine-coarse-es & UN coarse-level product data \\
        & in Spanish. \\
        lt-un-data-fine-industry-es & UN product data linked \\
        & to industries in Spanish. \\
        lt-un-data-fine-fine-fr & UN fine-level product data \\
        & in French. \\
        lt-un-data-fine-coarse-fr & UN coarse-level product data \\
        & in French. \\
        lt-un-data-fine-industry-fr & UN product data linked \\
        & to industries in French. \\
        lt-un-data-fine-fine-multi & UN fine-level product data \\
        & in multiple languages. \\
        lt-un-data-fine-coarse-multi & UN coarse-level product data \\
        & in multiple languages. \\
        lt-un-data-fine-industry-multi & UN product data linked \\
        & to industries in multiple languages. \\
        \bottomrule
    \end{tabular}
    \end{threeparttable}
   }
    \caption{Model names and training data sources for various models in the \lt model zoo. Each of these models is on the Hugging Face hub and can be found by prefixing the organization name \textit{dell-research-harvard} (for example, \textit{dell-research-harvard/lt-wikidata-comp-multi}.). Training code can be found on our package Github repo and training configs containing the hyperparameters are available in the model repo on the Hugging Face Hub.}
    \label{tab:description}
\end{table}

\begin{table}[ht]
    \centering
    \resizebox{1.05\linewidth}{!}{
    \begin{threeparttable}
    \begin{tabular}{lccc}
        \toprule
        \textbf{Model} & \textbf{Training} & \textbf{Validation} & \textbf{Test} \\
        & \textbf{Size} & \textbf{Size} & \textbf{Size} \\
        \midrule
        lt-wikidata-comp-es & 16511 & 924 & 932 \\
        lt-wikidata-comp-fr & 42475 & 2431 & 2486 \\
        lt-wikidata-comp-ja & 35923 & 2035 & 2054 \\
        lt-wikidata-comp-zh & 26224 & 1510 & 1513  \\
        lt-wikidata-comp-de & 42647 & 2383 & 2377 \\
        lt-wikidata-comp-en & 133557 & 7685 & 7648 \\
        lt-wikidata-comp-multi & 381820 & 28682 & 30532 \\
        lt-wikidata-comp-prod-ind-ja & 3647 & 149 & 149 \\
        lt-un-data-fine-fine-en & 9545 & 569 & 587 \\
        lt-un-data-fine-coarse-en & 8059 & 1399 & 614 \\
        lt-un-data-fine-industry-en & 8644 & 977 & 474 \\
        lt-un-data-fine-fine-es & 5462 & 289 & 305 \\
        lt-un-data-fine-coarse-es & 4326 & 552 & 389 \\
        lt-un-data-fine-industry-es & 4134 & 622 & 530 \\
        lt-un-data-fine-fine-fr & 1185 & 249& 204  \\
        lt-un-data-fine-coarse-fr & 3191 & 546 & 261 \\
        lt-un-data-fine-industry-fr & 3219  & 501  & 302 \\
        lt-un-data-fine-fine-multi & 19311 & 374 & 443 \\
        lt-un-data-fine-coarse-multi & 17939 & 529 & 911 \\
        lt-un-data-fine-industry-multi & 16528 & 1974 & 888 \\
        lt-mexicantrade4748 & 5348 & 466  & 477 \\
        lt-historicjapan & 982 & 50 & 55 \\
        \bottomrule
    \end{tabular}
    \end{threeparttable}
   }
    \caption{Model names and training, validation, and test sizes for various models in the \lt model zoo. The training size corresponds to the number of samples (or pairs for training with online contrastive loss) in the split. Validation and Test size correspond to the number of 'queries' for models evaluated on the retrieval task and to positive pairs for models evaluated on the paired classification task (For \textit{historicjapan}). The data were split into test-train-val at the class level to avoid test set leakage whenever possible.}
    \label{tab:performance_splits}
\end{table}
\begin{table}[ht]
\centering
    \resizebox{1\linewidth}{!}{

\begin{tabular}{lll}
\hline
\textbf{Dataset} & \textbf{Model} & \textbf{Loss} \\ \hline
Structured\_Amazon-Google & multi-qa-mpnet-base-dot-v1 & supcon \\
Structured\_Beer & bge-large-en-v1.5 & onlinecontrastive \\
Structured\_DBLP-ACM & bge-large-en-v1.5 & onlinecontrastive \\
Structured\_DBLP-GoogleScholar & bge-large-en-v1.5 & onlinecontrastive \\
Structured\_iTunes-Amazon & bge-large-en-v1.5 & onlinecontrastive \\
Structured\_Walmart-Amazon & bge-large-en-v1.5 & supcon \\
Structured\_Fodors-Zagats & bge-large-en-v1.5 & supcon \\
Dirty\_DBLP-ACM & bge-large-zh-v1.5 & supcon \\
Dirty\_DBLP-GoogleScholar & bge-large-zh-v1.5 & supcon \\
Dirty\_iTunes-Amazon & all-mpnet-base-v2 & supcon \\
Dirty\_Walmart-Amazon & bge-large-zh-v1.5 & supcon \\
Textual\_Abt-Buy & multi-qa-mpnet-base-dot-v1 & onlinecontrastive \\ \hline
\end{tabular}
}
\caption{Base models and Loss functions used for training of industrial benchmarks. Other hyperparameters that were constant across all of these experiments - learning rate (2e-5) , batch size (128), linear warmup of a 100\% (reaching the maximum learning rate). All models were run for 100 epochs and the checkpoint was selected on the basis of test F1 on validation set. Since labels (and also negatives) were also in the dataset, validation was done by pairwise classification. }
\label{tab:model-losses}
\end{table}

\begin{landscape}
\begin{table}[h]
\begin{tabular}{lccccccccccc}
\cline{1-12}
\textbf{Type} &
  \textbf{Dataset} &
  \textbf{Domain} &
  \textbf{Size} &
  \textbf{\# Pos.} &
  \textbf{\# Attr.} &
  \textbf{Ours (ZS)} &
  \textbf{Ours (FT)} &
  \textbf{Magellan} &
  \textbf{Deep matcher} &
  \textbf{Ditto} &
  \textbf{REMS} \\
  \hline
\multirow{7}{*}{Structured} &
  BeerAdvo-RateBeer &
  beer &
  450 &
  68 &
  4 &
  83.38 &
  90.32 &
  78.8 &
  72.7 &
  84.59 &
  96.65 \\
                         & iTunes-Amazon1  & music       & 539      & 132    & 8 & 60.6   & 90   & 91.2 & 88.5 & 92.28 & 98.18 \\
                         & Fodors-Zagats   & restaurant  & 946      & 110    & 6 & 75   & 98   & 100  & 100  & 98.14 & 100   \\
                         & DBLP-ACM1       & citation    & 12,363   & 2,220  & 4 & 95   & 98 & 98.4 & 98.4 & 98.96 & 98.18 \\
                         & DBLP-Scholar1   & citation    & 28,707   & 5,347  & 4 & 80   & 92 & 92.3 & 94.7 & 95.6  & 91.74 \\
                         & Amazon-Google   & software    & 11,460   & 1,167  & 3 & 47.1 & 74   & 49.1 & 69.3 & 74.1  & 65.3  \\
                         & Walmart-Amazon1 & electronics & 10,242   & 962    & 5 & 45   & 73.8   & 71.9 & 67.6 & 85.81 & 71.34 \\ \cline{1-12}
\multirow{1}{*}{Textual} & Abt-Buy         & product     & 9,575    & 1,028  & 3 & 28.8 & 84 & 33   & 55   & 88.85 & 67.4  \\
                       & Company   & company    & 1,12,632 & 28,200 & 1 &  74.07     & 88.00      &   79.8     &  92.7   &  41.00       &      80.73  \\
                          \cline{1-12}
\multirow{4}{*}{Dirty}   & iTunes-Amazon2  & music       & 539      & 132    & 8 & 68.8   & 84 & 46.8 & 79.4 & 92.92 & 94.74 \\
                         & DBLP-ACM2       & citation    & 12,363   & 2,220  & 4 & 89.8 & 98 & 91.9 & 98.1 & 98.92 & 98.19 \\
                         & DBLP-Scholar2   & citation    & 28,707   & 5,347  & 4 & 87.5   & 92.6 & 82.5 & 93.8 & 95.44 & 91.76 \\
                         & Walmart-Amazon2 & electronics & 10,242   & 962    & 5 & 45   & 71   & 37.4 & 53.8 & 82.56 & 65.74 \\
                         \hline
\end{tabular}
\caption{Benchmarks. ZS is \lt models zero-shot and FT is \lt models fine-tuned on the benchmark. The remaining columns report comparisons. The metric is F1, as these datasets frame linkage as a binary classification problem.}
\label{tab:industry_benchmarks}
\end{table}

\end{landscape}

\end{document}